\documentclass[tinyml]{acmart}
\setcopyright{rightsretained}
\copyrightyear{2025}
\renewcommand\footnotetextcopyrightpermission[1]{} 
\usepackage{graphicx}
\usepackage{subcaption}
\nointerlineskip
\begin{document}

\title{Enhancing Field-Oriented Control of Electric Drives with Tiny Neural Network Optimized for Micro-controllers}

\author{Martin Joel Mouk Elele}
\affiliation{%
  \institution{University of Pavia}
  \city{Pavia}
  \country{Italy}
}
\author{Danilo Pau}
\affiliation{%
  \institution{STMicroelectronics Inc.}
  \city{Milan}
  \country{Italy}
}
\author{Shixin Zhuang}
\affiliation{%
  \institution{Mathworks}
  \city{Boston}
  \country{USA}
}
\author{Tullio Facchinetti}
\affiliation{%
  \institution{University of Pavia}
  \city{Pavia}
  \country{Italy}
}


\begin{abstract}
 The deployment of neural networks on resource-constrained micro-controllers has gained momentum, driving many advancements in Tiny Neural Networks. This paper introduces a tiny feed-forward neural network, TinyFC, integrated into the Field-Oriented Control (FOC) of Permanent Magnet Synchronous Motors (PMSMs). Proportional-Integral (PI) controllers are widely used in FOC for their simplicity, although their limitations in handling nonlinear dynamics hinder precision. To address this issue, a lightweight 1,400 parameters TinyFC was devised to enhance the FOC performance while fitting into the computational and memory constraints of a micro-controller. Advanced optimization techniques, including pruning, hyperparameter tuning, and quantization to 8-bit integers, were applied to reduce the model's footprint while preserving the network effectiveness. Simulation results show the proposed approach significantly reduced overshoot by up to 87.5\%, with the pruned model achieving complete overshoot elimination, highlighting the potential of tiny neural networks in real-time motor control applications.
\end{abstract}

\maketitle

\section{Introduction}
\label{sec:intro}
Over the last decade, machine learning applications have expanded from traditional domains like computer vision to edge centric problems such as near-sensor data analytics and intelligent control in resource-constrained environments. EdgeAI, previously known as TinyML, addresses this challenge by enabling machine learning inference on ultra-low-power devices, such as micro-controllers, which prioritize power efficiency, low latency, and minimal memory use~\cite{10177729}. This paradigm supports localized intelligent computation, reducing cloud dependency while enhancing privacy and efficiency~\cite{9586232}. Tiny Neural Networks (TinyNN) have shown potential in fields requiring rapid and efficient monitoring and control with low computational overhead~\cite{10284551}. Still these cases have being focused on Internet of Things (IoT) and much less in domains such electric vehicles (EVs), with particular reference to electric motor control. 

Permanent Magnet Synchronous Motors (PMSMs), known for their high torque and power density, are pivotal in EVs applications such as automotive, industrial, naval and aeronautics, where compact size and precision control are essential~\cite{10126799}. PMSMs consist of a stator housing the windings and a rotor containing permanent magnets. The operational interaction between the stator’s rotating magnetic field and the rotor’s fixed magnetic field enables synchronization at synchronous speed~\cite{article}. Field-Oriented Control (FOC) is a widely adopted technique to obtain good control capability over the full torque and speed range for various motor types, including PMSM. 

Proportional-Integral (PI) controllers are simple and easy to implement but can be challenging to tune in situations where there are uncertainties and external disturbances. For example, in the case of PMSMs, electrical parameters such as resistance and inductance can change with wear and operating temperature, introducing uncertainties in system dynamics. To address the tuning challenges, advanced control techniques like $H_\infty$-synthesis~\cite{8217511} and Model Predictive Control (MPC) improve robustness but significantly increase control loop complexity~\cite{6734056}, causing unacceptable latencies on resource-constrained Micro-Controller Units (MCUs). 

\begin{figure}
	\centering
	\includegraphics[width=\columnwidth]{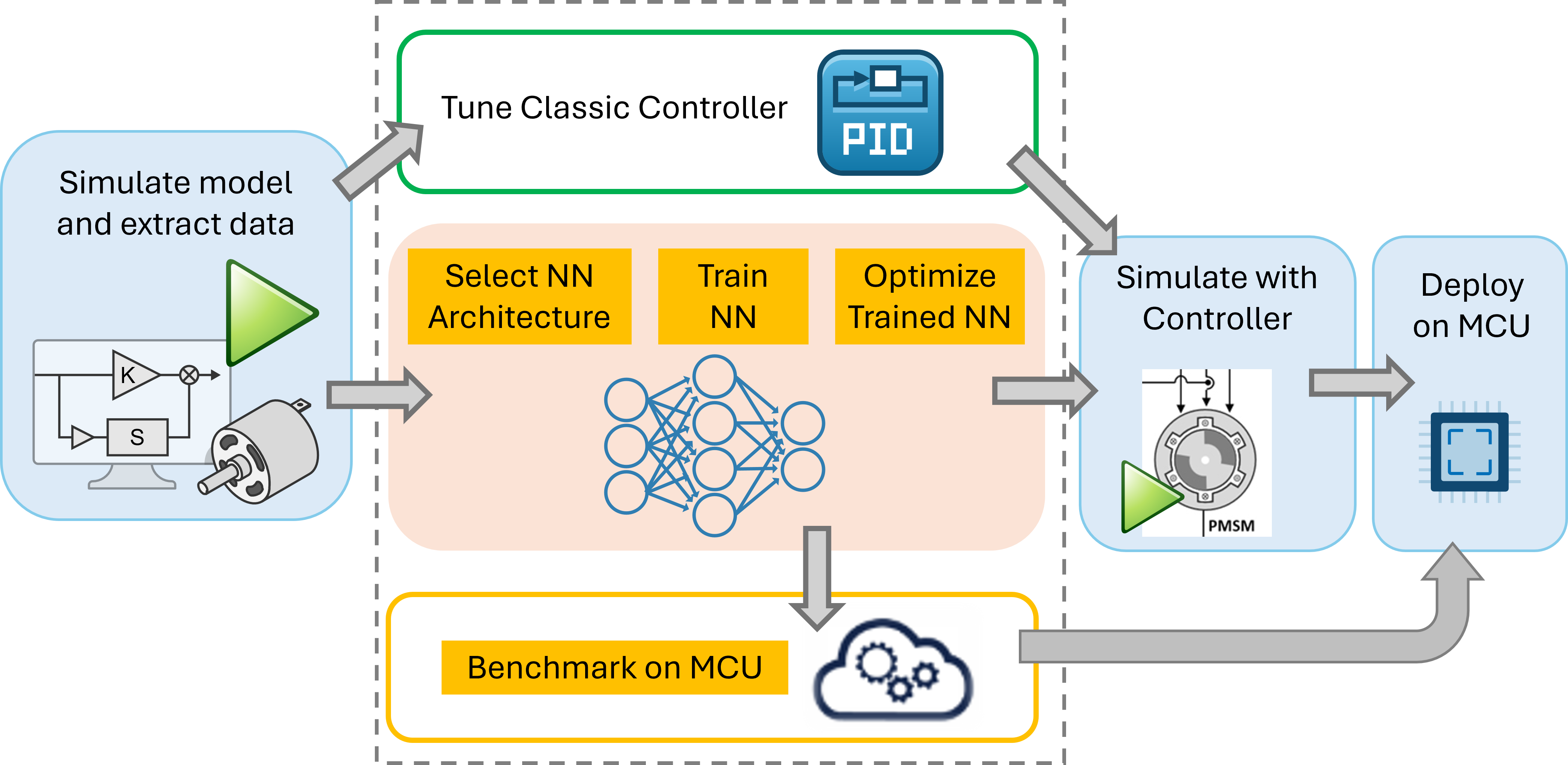}
	\caption[FrameworkDiagram]{Workflow diagram to deploy NN-augmented FOC \label{f:framework}}
\end{figure}

This work devises a tiny feed-forward Neural Network (NN), TinyFC, for FOC to enhance the responsiveness and precision of motor control systems. TinyFC, optimized for resource-constrained MCUs, provides improved performance while maintaining the strict power and memory constraints of embedded implementations. The design workflow of TinyFC is illustrated in Figure~\ref{f:framework}. Experimental results have been derived from two test cases specifically designed to test the controller in challenging operational scenarios.
The results demonstrated that this approach can meet the demands of modern EV control applications, showcasing the NN-augmented controllers to efficiently address real-world challenges.

The remainder of this paper is structured as follows. Section~\ref{sec:sota} reviews the related work in the domain of EdgeAI and the motor control applications, emphasizing the limitations of traditional controllers. Section~\ref{sec:systemMotorControl} presents the system approach to set up the simulation environment and focus on the challenging scenarios. Section~\ref{sec:Problem_def} and~\ref{sec:Optimization_tech} covers the dataset creation, NN training, optimization techniques for deployment on MCUs. Section~\ref{sec:experiment} shows and discusses the performance improvements achieved by the proposed neural network.

\section{Preliminaries and Related Work}
\label{sec:sota}

For PMSMs, FOC offers robust control over both torque and speed across the entire operational range. The implementation involves the transformation of stator currents from the stationary reference frame to the rotor flux reference frame, known as the $d$-$q$ reference frame~\cite{Krause2013}. Speed control and torque control are the most commonly used control modes of FOC. In the speed control mode, the motor controller follows a reference speed value and generates a torque reference for the torque control that forms an inner subsystem. The implementation requires real-time feedback of the currents and rotor position. 

PI-based FOCs are widely adopted for PMSM control since they simplify the system by regulating the direct axis current ($i_d$) to zero, enabling Maximum Torque Per Ampere (MTPA) operation. PI controllers adjust the motor’s input voltages in real-time based on feedback from current and speed sensors, ensuring the stator currents track the reference values~\cite{9771393}. However, PI controllers can struggle with system uncertainties and nonlinearity. To ensure robustness, PI-based FOC is augmented with additional algorithms like $H_\infty$-synthesis~\cite{8217511} and MPC.
In particular, the principle of MPC involves using a mathematical prediction model to forecast the future states of the controlled system within a prediction horizon. The controller then calculates a sequence of optimal control actions to track the desired reference trajectory, such as speed, while satisfying constraints, such as voltage and current. This enables better performance than PI-controllers by selecting optimal switching states for the inverter that modulates the power supply of the motor, while ensuring that the system stays within safe operational limits~\cite{10126799}.

Recent advancements in PMSM control have incorporated neural networks (NNs) to enhance traditional methods since they offered dynamic adaptability and precision, particularly useful for high-performance applications like EVs~\cite{9641002}. These NNs have been particularly effective in reducing torque ripple, which was a source of vibrations that affected ride quality and long-term mechanical durability~\cite{9641002}. The integration of Approximate Dynamic Programming (ADP) and the Levenberg-Marquardt algorithm had further accelerated the learning in these control systems~\cite{4596066}.

TinyNN, a subset of EdgeAI, are designed to deploy NNs on low-power, resource-constrained hardware while maintaining high accuracy. By leveraging the theoretical strengths of NNs, such as their universal approximation capability, TinyNN address the computational challenges faced by traditional NNs, making them suitable for embedded systems~\cite{HORNIK1989359}. TinyNN applications span on IoT domain like predictive maintenance, gesture recognition, and event detection, often in consumer devices such as smart home appliances, remote controllers, and wearables~\cite{10177729,lê2023efficientneuralnetworkstiny}.

\section{System Design}
\label{sec:systemMotorControl}

This section presents the design of the simulation framework utilized for developing the FOC for PMSM. The simulation environment is constructed in Simulink models with high fidelity. A dataset is collected to mimic the response of the motor when subjected to a wide set of input speed shapes, with specific attention to non-linear ones. A non-linear control based on a tiny NN is developed to match or outperform the behavior of a PI regulartor in certain scenarios. The simulation environment serves as a virtual testbed, enabling the iterative development and refinement of control algorithms. In the development process, benchmarking data are collected from the runtime resource profiling details from deployed C-inference models on the target MCU. The models, dataset and methods are made available on GitHub. \footnote{\url{https://github.com/heixiaopengyou/TINY-ML-for-FOC-of-PMSM-20092024}} 

\subsection{Setup and hardware}
\label{ssec:setup}

The motor model used in the FOC setup was based on the BR2804-1700KV-1 motor parameters,  controlled by using the X-NUCLEO-IHM07M1 inverter from the P-NUCLEO-IHM001 kit~\cite{st_nucleo_motor_control}. The BR2804-1700KV-1 motor used in the experiments has several key characteristics: it operates at a nominal voltage of $11.1V$ DC and can handle a maximum current of $5A$. It features 7 pole pairs and can achieve a maximum speed of $19000 rpm$.

\begin{figure}[H]
	\centering
	\includegraphics[width=\columnwidth]{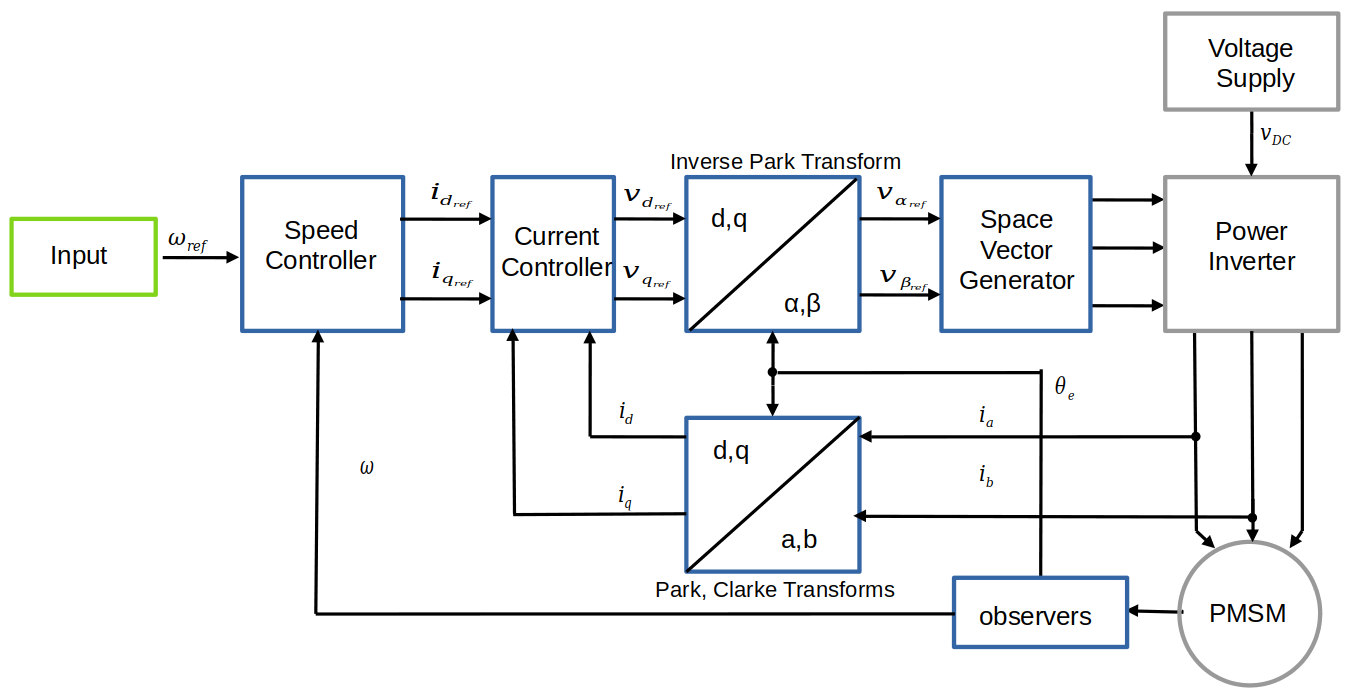}
	\caption[FOC setup modeled in MATLAB]{System model of input (in green), control (in blue) and motor/inverter (in gray) implemented in Simulink \label{f:FOC setup in Simulink}}
\end{figure}

The control system, as depicted in figure~\ref{f:FOC setup in Simulink}, consists of three subsystems: Input, FOC, and Motor/Inverter. The system model  is implemented in Simulink. This setup allows for a detailed examination of the motor's dynamic response to control inputs and environmental changes. The simulation environment serves as a virtual testbed, enabling the iterative development and refinement of control algorithms~\cite{NUSTES2023100479}.

\subsection{Test cases}
\label{ssec:usecase}

\begin{figure}[htbp]
    \centering
    \begin{subfigure}[t]{\columnwidth}
        \includegraphics[width=\columnwidth]{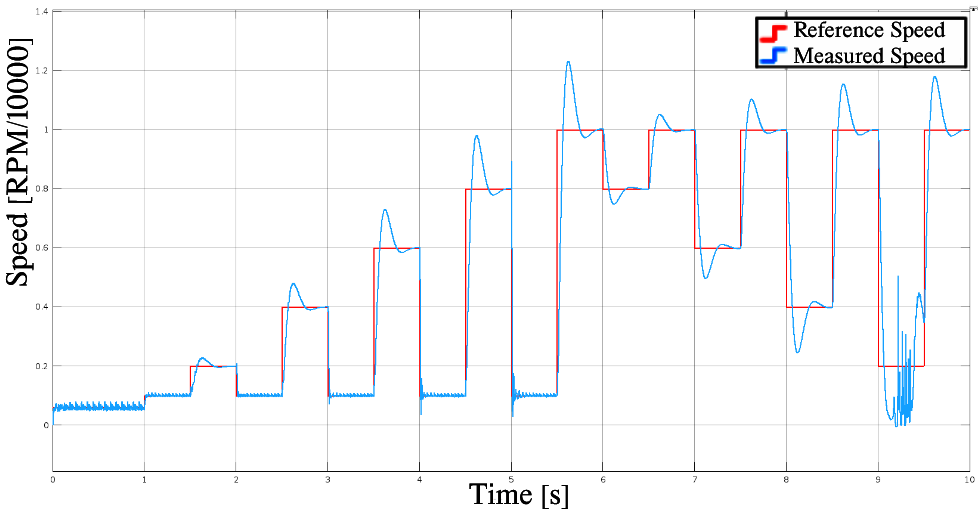} 
        \caption{Case 1: Reference speed of step signals}
        \label{f:Use case1}
    \end{subfigure}
    \begin{subfigure}[t]{\columnwidth}
        \includegraphics[width=\columnwidth]{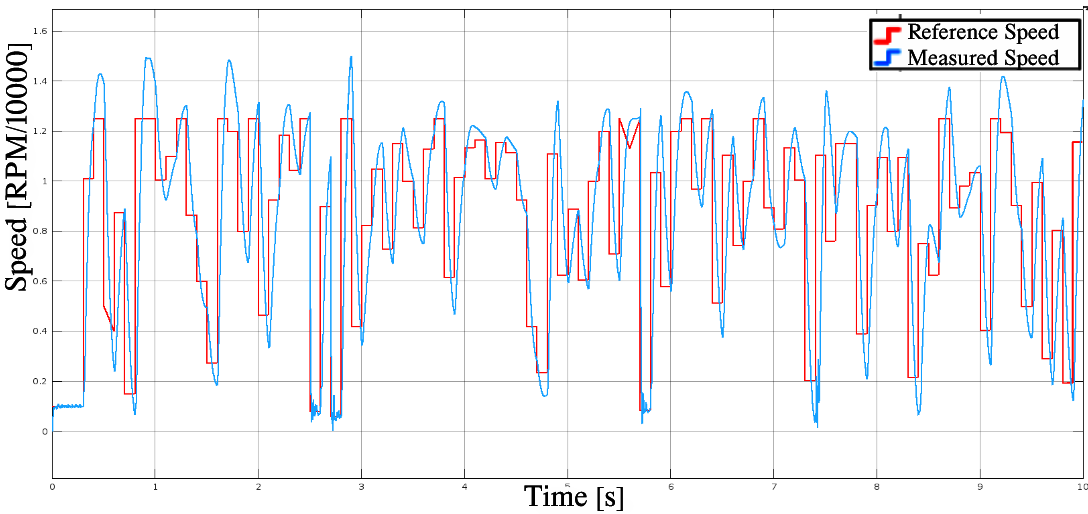} 
        \caption{Case 2: Reference speed of step and ramp signals}
        \label{f:Use case2}
    \end{subfigure}
    \caption{Measured speed collected from PI-based FOC}
\end{figure}

The test cases devised by this work are custom-designed signals aimed at testing the limits of the control loop and challenging the PI controller. These signals were specifically crafted to include multiple speed transitions of varying magnitudes and durations, as shown in figures~\ref{f:Use case1} and~\ref{f:Use case2}. 

Test case~1 was designed to test the controller’s response to a sequence of step signals with varying amplitudes~\cite{ogata-1987a}, and introduces a signal with 2 transitions per second~\cite{NUSTES2023109002}. Tracking results using conventional PI-based FOC are shown in figure~\ref{f:Use case1}.

In test case~2, the input signal is a combination of step and ramp signals~\cite{ogata-1987a} with 10 speed transitions per second, designed to assess the controller's ability to manage both abrupt changes and gradual variations in speed.
For test case~2, the tracking accuracy of the PI controller is shown in figure~\ref{f:Use case2}.

\section{Training of the Neural Network} 
\label{sec:Problem_def}

Neural networks are trained to classify abstract patterns or predict outcome without having experience or existing knowledge to draw from.
Our dataset simulates the behavior of a PMSM motor controlled using the FOC scheme, and it is made by the parameters reported in table~\ref{t:dataset}.

\begin{table}[htbp]
    \centering
    \begin{tabular}{lcc}
    \toprule
    \textbf{Notation} & \textbf{Parameter} &  \textbf{I/O} \\
    \midrule
         $\omega^{ref}$ & Reference Speed &  Input \\
         $\omega^{meas}$ & Measured Speed &  Input \\
         $i_q^{PI}$ & PI-predicted Quadrature Current &  Input\\
         $\Delta i_q^{TinyFC}$ & Current Compensation & Output\\
    \bottomrule
    \end{tabular}
    \caption{Parameters composing the dataset used in the training of the NN model\label{t:dataset}}
    
\end{table}

%
%
PI-based FOC assumes precise predictions in the speed control loop in normal operation of PMSM. However, uncertainties and external disturbances, such as operating temperature or system dynamics, can lead to performance degradation for PI controllers~\cite{inbook}. In both cases described in section~\ref{ssec:usecase}, the controller performed poorly. This happened when the calculated reference quadrature current generated by the speed PI controller contained deviations (errors) for most of the time steps. When the system dynamic changes, the PI controller demonstrates sluggish responses, overshoots, stability issues and tracking errors.
To improve the quality of the ground truth signal, the prediction made by the PI controller in the speed control loop was adjusted to eliminate the sections containing overshoots.
The remainder of this section discusses the impact of how the adjusting approaches affect the signals in the test cases.

In test case 1, simulations show that the PI controller struggle to quickly adapt to abrupt changes in the reference speed, leading to poor dynamic performances and sluggish responses. 
Moreover, it produces a significant deviation, as can be seen from the maximum deviation for PI-only control in table~\ref{t:PIvsTinyNN}, and longer settling times, impacting the precision and stability of motor control. The PI controller exhibits oscillations before stabilizing, causing abnormal spikes in the predicted quadrature current.
By analyzing the reference quadrature current and response speed, 
it was possible to identify when the quadrature current exceeded the desired range.
This range was defined using a threshold $C$, representing the maximum acceptable value within the response interval~\cite{1053958}.
The modified signal was thus determined by the following equation: 

 \begin{equation}\label{Threshold_formulation}
     x^{adj}(t) = min\{|x(t)|, C\}
 \end{equation}
 where $x^{adj}(t)$ is the saturated signal.
 
\begin{figure}
	\centering
	\includegraphics[width=0.49\columnwidth]{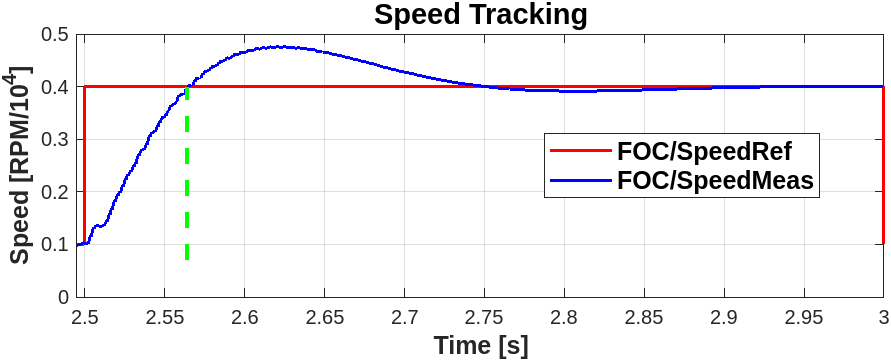}
	\includegraphics[width=0.49\columnwidth]{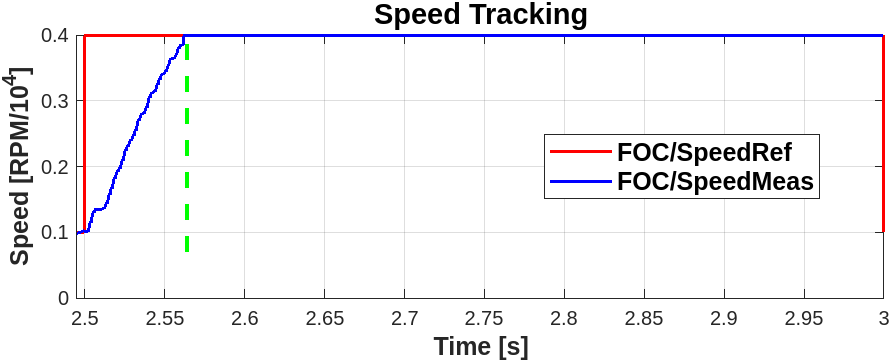}
    
	\includegraphics[width=0.49\columnwidth]{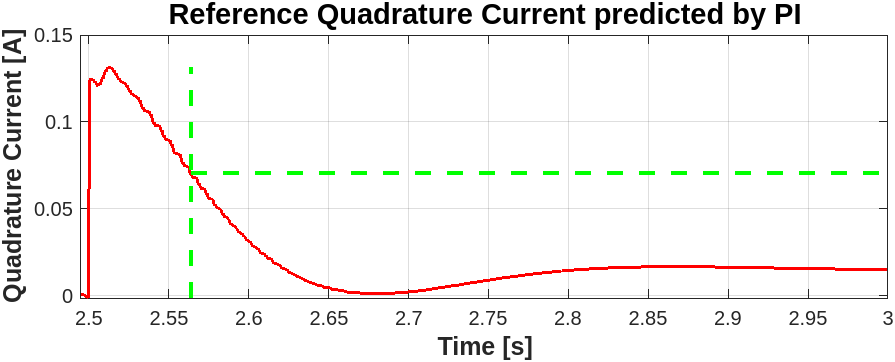}
	\includegraphics[width=0.49\columnwidth]{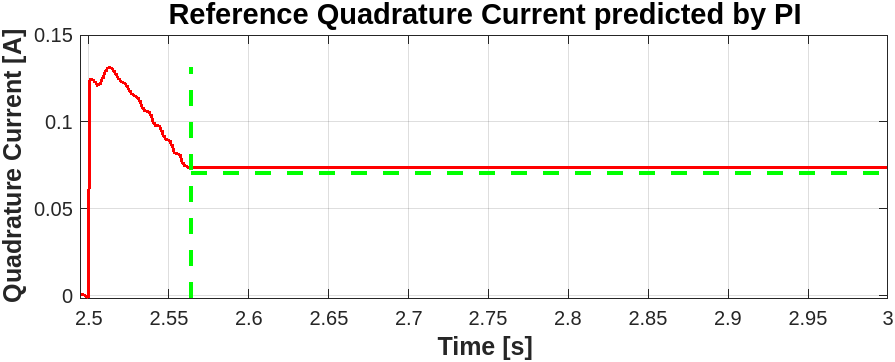}
	\caption[Threshold Case 1]{Quadrature current from PI-controller simulation adjusted using speed observations threshold\label{f:Applying a threshold}}
\end{figure}

If the response signal $x(t)$ exceeds the threshold $C$ in magnitude, the signal saturates to $C$ as shown in Figure~\ref{f:Applying a threshold}. 
 
\begin{figure}
	\centering
	\includegraphics[width=\columnwidth]{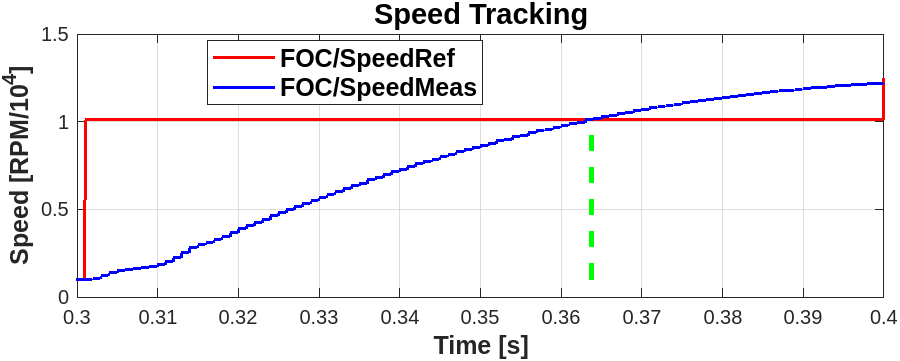}
	\includegraphics[width=\columnwidth]{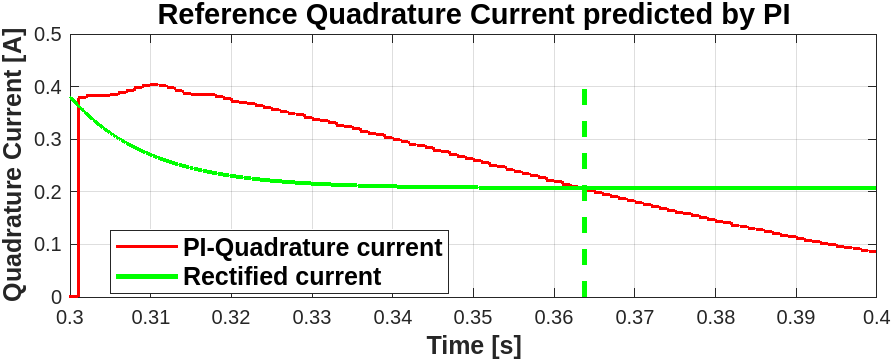}

	\caption[Desired Signal path]{Quadrature current adjusted based on capping and rectification of measured speed\label{f:Current cap and adjustment}}
\end{figure}

In test case 2, simulations show that with an increased speed transition frequency, although the PI controller followed the overall speed trend, it significantly fails to stabilize around the desired speed (for each interval). The instability of the response signal, i.e., the measured speed, posed greater challenges, as shorter intervals of steady-state behavior reduced reliability.
The proposed approach involves rectifying the reference quadrature current $i_q^{PI}$.
By analyzing intervals where the measured speed matches the reference speed, the initial and final in-range values of the quadrature current are used to generate a rectified signal.
This rectification employs an exponential decay, or growth curve, aligning the initial and final in-range values of the quadrature current with the start and end points of the function.
The corrected signal ensures that the reference quadrature current more closely mimicks an ideal response, remaining within the desired range throughout each interval as shown in Figure~\ref{f:Current cap and adjustment}.
The corrected signal per interval hence becomes:

\begin{equation}\label{current_decay_setup}
    x^{adj}(t) = x^{final}(t) + (x^{initial}(t) - x^{final}(t))e^{-(t/\tau)}
\end{equation}
where $\tau$ is the decay constant ensuring that the current response is not too fast nor too slow, avoiding sharp oscillations in the response.

As a result of the previous considerations, the signal used as ground truth for training the TinyFC NN is $\Delta i_q^{GT}$ defined as:

\begin{equation}
\Delta i_q^{GT} = x(t) - x^{adj}(t)
\end{equation}

\section{Neural network design}
\label{sec:Optimization_tech}

\begin{figure}[H]
	\includegraphics[width=\columnwidth]{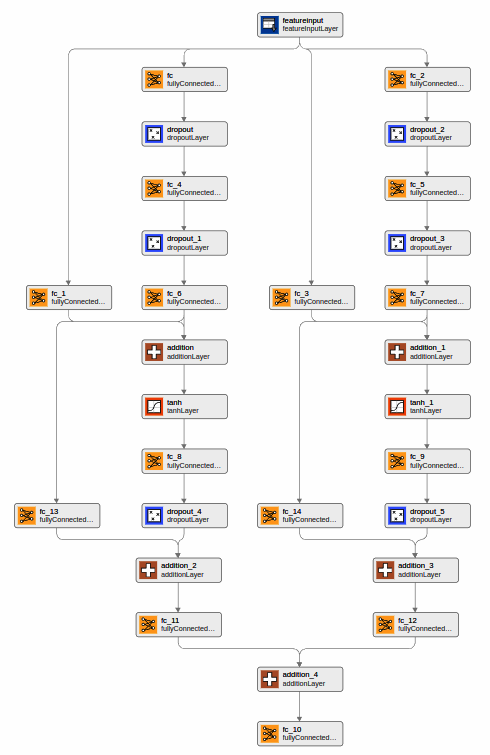}
	\caption[Network model topology as shown by MATLAB]{Network model topology as shown by MATLAB \label{f:tinyNN_topology}}
\end{figure}

The TinyFC proposed in this work, whose architecture is shown in Figure~\ref{f:tinyNN_topology}, is a $1,400$ parameter model prior deployed using the ST Edge AI Developer Cloud~\cite{st_edge_ai_developer_cloud} to ensure deployability to the MCUs.

The model inputs/outputs are reported in table~\ref{t:dataset}.
The output of the TinyFC NN consists in the adjustment value $\Delta{i_q}$ for the PI's prediction that is used to correct the reference signal $i_q^{ref(PI)}$ based on the following equation:

\begin{equation}\label{quadrature_correction}
    i_q^{adj} = i_q^{ref(PI)} + \Delta{i_q}^{TinyFC}
\end{equation}
where $i_q^{adj}$ is the actual signal used for controlling the motor.

The model consists of branches of fully connected (FC) layers, combining $2$ main-branches of $5$ FC layers, and $2$ residual FC layers per branch.
The outputs of the two branches are merged using $tanh$ activations, constraining them to the range [-1, 1].
Training data comprised a 300001$\times$4 sequence per test case, derived from the Simulink PI-based FOC model. Each 10-second test case sampled at $T_s = 3.3333 \times 10^{-5}$, corresponding to the PWM switching frequency, yielding $\sim 300,000$ samples.
The NN was trained using data from the test case~1, and fine-tuned on data from test case~2. For training, the dataset was split into a training, validation and test set in the ratio  8:1:1, in order to validate the model on unseen data.  Once trained, the TinyFC was integrated into the FOC speed control unit, and its corrective performance was evaluated.

A key challenge of integrating a NN in an FOC loop is its inference time, as the PWM operating at 30 kHz requires inputs every 33.33 $\mu$s. If the NN's inference exceeds this time, system delays and performance issues may occur. To mitigate this, the NN was optimized by reducing the number of parameters, minimizing both computational and memory demands.
Two alternative optimization approaches have been considered: hyperparameters optimization and pruning.
In both cases, $8$-bit quantization was applied to further reduce the memory footprint.

\subsection{Hyperparameters optimization}
\label{ssec:hpo}

Hyperparameters govern model performance.
Therefore, hyperparameter optimization (HPO) is essential for optimal model performance and complexity.
Using Bayesian Optimization (BO), which efficiently balances exploration and exploitation~\cite{mathworks_bayesian_optimization,10467454}, the model's hyperparameters were fine-tuned to achieve optimal predictions for speed monitoring in PMSM FOC. This process identified the best-performing configuration for deployment.

\subsection{Pruning}
\label{ssec:prune}

Projection pruning, leveraging Principal Component Analysis (PCA), was applied to compress the model by retaining only the most influential neurons based on variance in activations~\cite{1244591}.
This reduced the model size by eliminating redundant connections while preserving accuracy.
PCA-based pruning identified key neurons by solving an eigenvalue problem on the covariance matrix of standardized neuron activations~\cite{Chen2001}.

\subsection{Quantization}
\label{ssec:quant}

An additional step in the optimization process was quantization.
All the models, i.e., the trained, HPO and pruned models, use a representation based on $32$-bits floating point numbers.
An integer $8$-bit quantization was thus applied to further reduce the memory footprint of the model.  
Post training quantization (PTQ) was adopted by using ST Edge AI Developer Cloud~\cite{st_edge_ai_developer_cloud}.

\section{Experimental Results}
\label{sec:experiment}

To evaluate the performance of the devised TinyFC in the FOC loop, three sets of metrics are evaluated: 
\begin{itemize}
    \item Neural network metrics, i.e. accuracy
    \item Control loop metrics, i.e. deviation statistics
    \item Deployability metrics, i.e. inference time on MCU
\end{itemize}

\subsection{Accuracy evaluation}
\label{ssec:metric1}

To evaluate the predictive accuracy, comparing the NN's predictions to ground truth values derived from the modified quadrature current, the Mean Squared Error (MSE) is evaluated:

\begin{equation}
    \text{MSE} = \frac{1}{n} \sum_{i=1}^{n} \left(\Delta i_q^{GT} - \Delta i_q^{TinyFC}\right)^2
\end{equation}
where $\Delta i_q^{GT}$ is the ground truth (PI prediction error), and $\Delta i_q^{TinyFC}$ is the NN's prediction.
Model complexity, quantified by the number of learnable parameters, also plays a critical role, as it directly impacts computational efficiency and deployability.

Table~\ref{t:Models_MSE} reports the evaluation of the MSE for the two considered test cases and the different configurations of optimized NNs.

\begin{table}
	\centering
	\small
	\begin{tabular}{c|lrr}
		\hline
		\textbf{Test} &\textbf{Model} & \textbf{Number of} & \textbf{MSE} \\
		\textbf{Case} & & \textbf{Parameters} & \textbf{(\%)} \\
		\hline
		& \textbf{TinyFC} & $1,400$ & $4.09$ \\
		1 & \textbf{HPO TinyFC} & $670$ & $0.31$ \\
		& \textbf{Pruned TinyFC} & $873$ & $9.19$ \\
        \hline
        & \textbf{TinyFC} & $1,400$ & $0.62$ \\
		2 & \textbf{HPO TinyFC} & $340$ & $0.62$ \\
		& \textbf{Pruned TinyFC} & $600$ & $30.45$ \\
		\hline
	\end{tabular}
	\caption[Models MSE case1&2]{Mean Squared Error of the trained NNs\label{t:Models_MSE}}
\end{table}
The optimization procedure demonstrated a significant reduction in the model's parameters, achieving up to a 75.7\% reduction in test case~2 with the HPO model.
However, it is notable that while combining pruning with HPO could yield even fewer parameters, the HPO model did not perform satisfactorily within the FOC loop during this study (this case is not shown in this paper).
Consequently, pruning was applied exclusively to the 1.4K-parameter TinyFC model for both test cases.

\subsection{Control loop evaluation}

The performance of the AI-augmented FOC was evaluated by monitoring overshoots and deviations during the simulations.
The maximum overshoot represents the highest peak above the intended steady-state response, while deviation measures the difference between the reference and the controller's output. Maximum deviation identifies challenging control regions needing further optimization, while average deviation assesses overall tracking consistency~\cite{ogata-1987a}.

The designed test cases expose the limitations with respect to FOC evaluation metrics. In test case~1, a noticeable overshoot occurs after each speed increase. In test case~2, the PI controller fails to stabilize.
This latter effect is noticeable by comparing the performance of the PI controller in the two test cases, as can be seen in table~\ref{t:PIvsTinyNN}: the signal diverges significantly from the reference speed, causing the maximum deviation to rise from $0.81$ in the first case to $1.21$, corresponding to a 49\% increase.
As observed in Figure~\ref{f:Use case2}, there is a considerable delay between the reference signal’s transition and the response of the measured speed, implying a lag in the system’s ability to react instantaneously.

The TinyFC with 1,400 parameters was able to reduce the deviation in the control system between the reference speed signal and the measured speed signal, outperforming the PI-based FOC, with a maximum overshoot correction of up to $87.5\%$ in test case 1, and $68\%$ in test case~2, as shown in table~\ref{t:PIvsTinyNN}.
This comparison of the performance highlights the capacity on the NN to detect the error in the PI's prediction and correct it.

\begin{table}
	\centering
	\small
	\begin{tabular}{c|lccc}
		\hline
		\textbf{Test} & \textbf{Controller} & \textbf{Max} & \textbf{Average} & \textbf{Max} \\
		\textbf{Case} & & \textbf{Deviation} & \textbf{Deviation} & \textbf{Overshoot} \\
		\hline
		& \textbf{PI} & 0.81 & 0.05 & 0.24\\
		1 & \textbf{PI + TinyFC} & 0.89 & 0.02 & 0.03\\
		& \textbf{\% Change} & +9 & -60 & -87.5 \\
		\hline
        & \textbf{PI} & 1.21 & 0.18 & 0.25\\
		2 & \textbf{PI + TinyFC} & 1.19 & 0.15 & 0.08\\
		& \textbf{\% Change} & -1.65 & -16.7 & -68 \\
		\hline
	\end{tabular}
	\caption[PI vs PI + TinyFC]{Performance comparison between PI and the TinyFC NN in terms of max and average deviation and max overshoot\label{t:PIvsTinyNN}}
\end{table}

\begin{table}
	\centering
	\small
	\begin{tabular}{c|lccc}
		\hline
		\textbf{Test} & \textbf{Model} & \textbf{Max} & \textbf{Average} & \textbf{Max} \\
		\textbf{Case} & & \textbf{Deviation} & \textbf{Deviation} & \textbf{Overshoot} \\
		
        \hline
		& \textbf{TinyFC} & 0.89 & 0.02 & 0.03\\
		1 & \textbf{HPO TinyFC} & 0.896 & 0.0316 & 0.06\\
        & \textbf{Pruned TinyFC} & 0.9 & 0.02 & $\sim 0$\\
        \hline
        & \textbf{TinyFC} & 1.19 & 0.15 & 0.08\\
		2 & \textbf{HPO TinyFC} & 1.23 & 0.15 & 1.24\\
        & \textbf{Pruned TinyFC} & 1.20 & 0.16 & 0.03\\
		\hline
	\end{tabular}
	\caption[Model Comparison]{Performance Comparison of Different Models\label{t:Model_Comp}}
\end{table}

As explained in subsection~\ref{ssec:hpo}, the TinyFC is optimized to obtain a lightweight model with a similar accuracy as the originally trained model.
Table~\ref{t:Model_Comp} shows that, inserting the HPO model in the control loop actually increases the overshoot by up to 100\%, with the HPO model unable to properly grasp the relationship between its inputs and its expected output. The pruned model further enhances the performance of the FOC, with $\sim$100\% reduction in the overshoot (test case~1).
The pruned model considerably reduces the overshoot in test case~2, despite slightly increasing the deviation from the reference signal in both cases.
\begin{figure}
	\centering
	\includegraphics[width=0.45\textwidth]{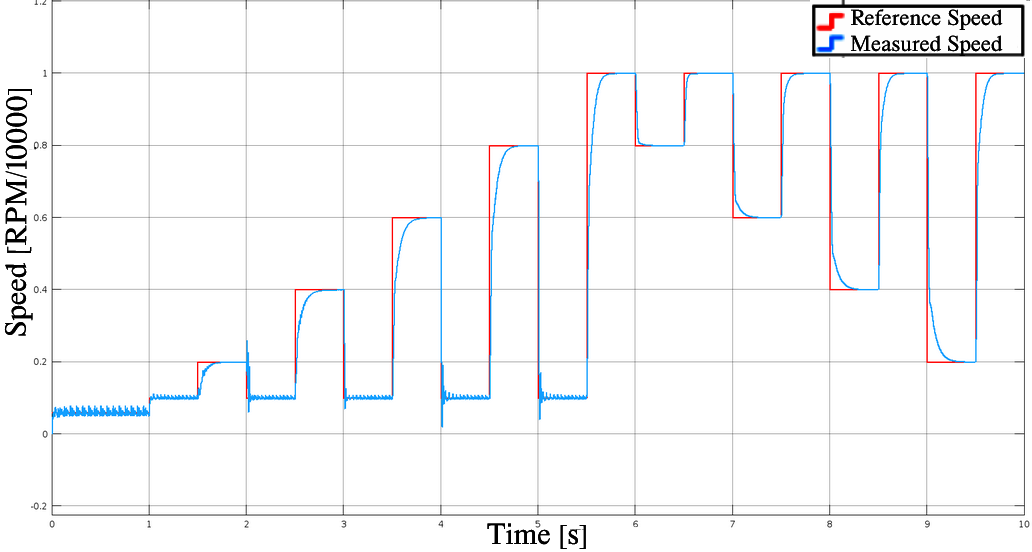}
	\caption[Performance with Pruned model(case 1)]{Output of pruned NN of test case 1 \label{f: Augmented Control loop with Pruned Model (case 1)}}
\end{figure}
Despite the observed improvement in test case~2, the augmented FOC does not perform as well as in the test case~1, exposing some of the limitations of this approach.

\subsection{Deployability on the MCU}
\label{ssub:deploy}

Further validation is considered by deploying the models in the ST Edge AI Developer Cloud, an online farm where models can be tested on real MCUs.

The considered metrics for assessing the suitability of the proposed approach for the deployment on an MCU are: 

\begin{itemize}
    \item the number of multiplications and accumulations (MACC) required to run the model; 
    \item the flash memory footprint to permanently store the application firmware; 
    \item the RAM footprint, which is the dynamic memory needed to store intermediate results while running the model;
    \item the inference time required by the model to generate an output after an input is provided during real-time operations.
\end{itemize}

This work considered the deployability of the optimized TinyFC models on an off-the-shelf MCU, namely NUCLEO-G474RE, which is widely used for PMSM control in industrial and automotive applications, and FOC in particular.
The primary objective was to ensure that the trained and optimized models meet the MCU embedded memory and real-time processing requirements for FOC.

\begin{table}
	\centering
    \small
	\begin{tabular}{p{0.72cm}|p{1cm}p{0.8cm}p{1.5cm}p{1.1cm}p{0.8cm}}
		\hline
		\textbf{Test Case} & \textbf{Model} & \textbf{MACC} & \textbf{FLASH (KiB)} & \textbf{RAM (KiB)} & \textbf{Time ($\mu$s)} \\
		\hline
		&TinyFC & 1620 & Weights: 5.68 Lib: 15 & Act: 0.199 Lib: 6 &207.4\\
		1 & HPO & 764 & Weights: 2.61 Lib: 15 & Act: 0.152 Lib: 6 & 144.8\\
		& Pruned & 1034 & Weights: 3.28 Library: 20 & Act: 0.148 Lib: 10 & 232.2\\
		& Quantized Pruned & 910 & Weights: 1.37 Lib: 32 & Act: 0.383 Lib: 13 & 371.7\\
        \hline
        &TinyFC & 1620 & Weights: 5.68 Lib: 15 & Act: 0.199 Lib: 6 & 207.4 \\
		2 & HPO & 470 & Weights: 1.33 Lib: 15 & Act: 0.105 Lib: 6 & 127.6 \\
		& Pruned & 760 & Weights: 2.26 Lib: 20 & Act: 0.145 Lib: 10 & 215.2 \\
		&Quantized Pruned & 640 & Weights: 1.07 Lib: 32 & Act: 0.360 Lib: 13 & 361.4 \\
		\hline
	\end{tabular}
	\caption[Model Performances on NUCLEO-G474RE]{Neural Network characteristics on NUCLEO-G474RE}
	\label{t:NN on NUCLEO-G474RE}
\end{table}

The NUCLEO-G474RE development board, with STM32G4 MCU running at 170MHz, is specifically designed to strike a balance between processing power and energy efficiency, making it suitable for embedded control applications. With embedded 512 KiB of flash memory and 128 KiB of RAM~\cite{st_nucleo_64_stm32g474re}, the NUCLEO-G474RE can handle moderately complex NN models.

Table~\ref{t:NN on NUCLEO-G474RE} shows the comparative results of the different versions of the NN devised by this work, including the quantized version of the pruned model, which offered the best error correction in the FOC setup in Simulink. It is observable that, despite the pruned model having a lower MACC number and less parameters than the original model with $1,400$ parameters, its inference time is surprisingly greater than that of the latter model, for both cases. 

\subsection{Discussion}

The challenge faced by the optimized models (HPO model in particular) implied that trimming less important weights or obtaining a model with the smallest MSE was not enough to guarantee the best performance in the FOC setup. 

This was because MSE as the evaluation metric for the NN during training and validation was not enough to capture how a specific change in quadrature current can affect the stability of the system.
Hence, a more robust indicator was required, which would take into account the physics (physical loss) related to the system for each prediction of the NN. Overall, the results from the TinyFC and its pruned version clearly showcase the benefits of using a NN in the control loop, which considerably reduces the deviations and overshoot that are generated due to the limitations of the PI controller.

The PI-based FOC, without the TinyFC, shown signs of underdamping, due to the presence of the overshoots and oscillations.
The non-optimized TinyFC does reduce the oscillations, which could be interpreted as adjustments in damping~\cite{atkins2013dictionary}, since the oscillations in the speed response were reduced.
The pruned model further reduced them.



\section{Conclusions}
\label{conclusion}

This research demonstrated the potential of TinyNN to improve FOC by minimizing overshoots and tracking errors in PI-based control. By leveraging reference speed, measured speed, and the PI controller's output, the proposed TinyFC model enhanced speed tracking and reduced steady-state errors.
However, optimizing TinyFC inference for MCU timing remains a challenge. Limitations in using MSE suggest exploring physics-informed neural networks (PINNs) to better model input dependencies through physics-based loss functions~\cite{TRASK2022110969,9593574}. 
While effective, the approach depends on the PI controller for reference adjustments, limiting TinyFC to a supportive role.

Future works could focus on developing a fully data-driven TinyNN model to replace the PI controllers, achieving a unified control system free from traditional methods.

\newpage
\bibliographystyle{plain}
\bibliography{acmart}
\end{document}